International Journal of Engineering Trends and Technology       Volume 71 Issue 3, 120-129, March 2023
ISSN: 2231 – 5381 / https://doi.org/10.14445/22315381/IJETT-V71I3P213       © 2023 Seventh Sense Research Group®*Review Article*

# Alzheimer's Disease Diagnosis using Machine Learning: A Review

Nair Bini Balakrishnan[1], P.S. Sreeja[2], Jisha Jose Panackal[3]

[1,2]*Department of Computer Science, Hindustan Institute of Technology and Science, Chennai, India.*
[3]*Department of Computer Science, Sacred Heart College, Kerala, India.*

[1]*Corresponding Author : biniremesh@gmail.com*Received: 25 November 2022     Revised: 11 March 2023     Accepted: 17 March 2023     Published: 25 March 2023***Abstract*** *- Alzheimer's Disease (AD) is an acute neuro disease that degenerates the brain cells and thus leads to memory loss progressively. It is a fatal brain disease that mostly affects the elderly. It steers the decline of cognitive and biological functions of the brain and shrinks the brain successively, which in turn is known as Atrophy. For an accurate diagnosis of Alzheimer's disease, cutting-edge methods like machine learning are essential. Recently, machine learning has gained a lot of attention and popularity in the medical industry. As the illness progresses, those with Alzheimer's have a far more difficult time doing even the most basic tasks, and in the worst case, their brain completely stops functioning. A person's likelihood of having early-stage Alzheimer's disease may be determined using the ML method. In this analysis, papers on Alzheimer's disease diagnosis based on deep learning techniques and reinforcement learning between 2008 and 2023 found in google scholar were studied—sixty relevant papers obtained after the search was considered for this study. These papers were analysed based on the biomarkers of AD and the machine-learning techniques used. The analysis shows that deep learning methods have an immense ability to extract features and classify AD with good accuracy. The DRL methods have not been used much in the field of image processing. The comparison results of deep learning and reinforcement learning illustrate that the scope of Deep Reinforcement Learning (DRL) in dementia detection needs to be explored.*

***Keywords -*** *Alzheimer's Disease, Deep Learning, Convolutional Neural Network, Recurrent Neural Network, Deep Neural Network.*## 1. Introduction

Among many types of Dementia, Alzheimer's Disease (AD) is the most common, which could cause serious damage to memory. AD is prevaded by localized brain atrophies. It deteriorates the key biological functions of neurons, such as communication, metabolism, repair, re-modelling, and regeneration. According to the latest studies, an estimated 6.2 million Americans aged 65 and older will be diagnosed with AD in 2021.[1] The cognitive abilities affected by AD include thinking, reasoning, and remembering. The preambulatory phase of AD is Mild Cognitive Impairment (MCI), which leads to cumulative memory loss and poor behavioural issues. According to the report in [2], about 15-20% of older persons (aged 65 or above) have MCI and 30-40% of them develop AD within 5-6 years. In most cases, the time period for this conversion is found to be 18 months, though it may vary from 6 to 36 months. AD is 6 percent more commonly found in females than males over the age of 65. [43] A conclusive diagnosis of AD can only be done with the tissue pathology of Beta-Amyloid plaques and neurofibrillary entangle during autopsy. The plaques are developed between the brain's neurons by collecting Beta-Amyloid protein, which disrupts neuron functions. On the other hand, the neurofibrillary entangles are abnormal accumulations of a protein called Tau. It is obvious that some benchmark other than the autopsy is required to confirm AD. In 2011, the National Institute on Aging and the Alzheimer's Association divided the biomarkers into two categories: (1) the biomarkers based on PET imaging and (2) the biomarkers based on Cerebrospinal Fluid as well as MRI.[3] The criteria for AD are being updated every 3 – 4 years so that new knowledge to track AD is inculcated.[4] Till date, no complete cure for AD has been discovered. Hence it becomes essential to detect this disease in its early stage itself. The average lifespan of an Alzheimer's patient is from 3 to 9 years.[41] Early diagnosis of AD can help to slow down the rate of speeding up of AD symptoms in the patients, and thus, worst dementia conditions can be deaccelerated. The crucial step is to detect the MCI conversion into AD rather than the classification of AD patients. The currently available computer-aided mechanisms cannot replace the medical experts in the field but can perform as an aiding tool to enhance their clinical diagnosis. Some review studies on AD diagnosis using machine learning techniques have been published. These studies are based on different types of biomarkers, feature selection methods, and single-modality and multi-modality approaches.

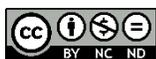
This is an open access article under the CC BY-NC-ND license (http://creativecommons.org/licenses/by-nc-nd/4.0/)

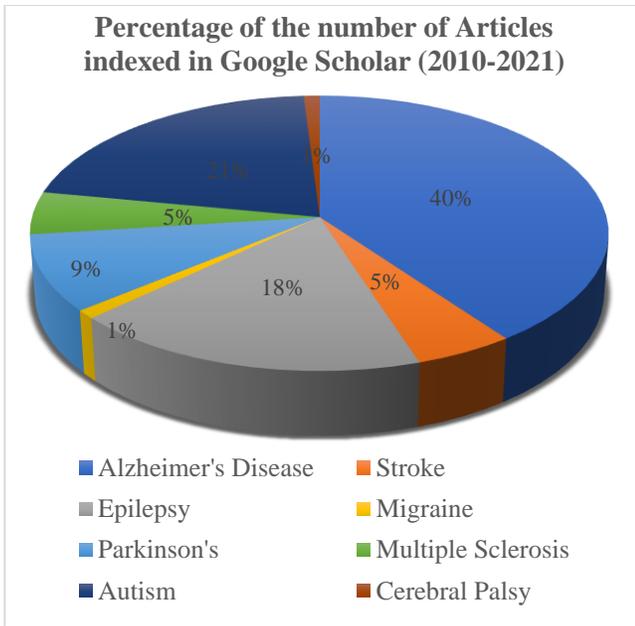

**Fig. 1 Articles on different neurological disorders published in Google Scholar (2010-2021)**

The interest of researchers in diagnosing AD has been increasing every year compared to other neurological disorders, as shown in Figure 1. Though various Machine Learning models have been developed for AD diagnosis, the literature review conducted for this study indicates that dedicated ML methods combining the strength of both deep learning and reinforcement learning for early AD diagnosis still need to be investigated. This study aims to fill the above-mentioned research gap.

### 1.1. Signs and symptoms
#### 1.1.1. Early Stage
Alzheimer's disease is a neurological disorder that gets worse over time and has three stages. These three stages, especially the early stage, are hard to spot and often confused with aging or stress. Problems with Alzheimer's disease treatments may not become apparent for as long as eight years, depending on the type of tests the doctor uses. Daily activities become increasingly challenging for the affected individual in the beginning stages. The most common sign is short-term memory loss, which makes it hard to remember the recent things a person has learnt and creates difficulty in learning new things. People affected with AD may have a lack of energy and a general sense of sadness in the beginning.[44] The most prevalent symptom of sickness is apathy or a general lack of care or interest. Mild cognitive impairment is a neurological disorder that may impede a person's ability to do daily tasks, learn new information, and make sense of the environment (MCI). Memory impairment (MCI) is characterized by a decline in short-term memory and is seen in the early stages of Alzheimer's disease. Studies have shown that a person with amnestic MCI has a 90% chance of getting Alzheimer's disease. The main signs of a language problem are having trouble dealing with hard words. AD patients face trouble in doing activities such as reading, writing, dressing, arranging, or planning things, but they cannot identify them.[50] A person with Alzheimer's disease can still do daily tasks with the help of a dependent person.[45]

#### 1.1.2. Middle Stage
AD patients have trouble conversing when they cannot remember words, which causes them to use words incorrectly. People find it difficult to juggle multiple responsibilities during this time. Another individual loses the capacity to recognize close relatives. An individual's ability to recognize his/her close relatives deteriorates over time. The individual's way of performing a task will change, making it more difficult for them to give their full mental attention. 30% of people with Alzheimer's disease tend to think that the identity of other people, places, and things has changed. Gradually, the person develops trouble figuring out who they are, which makes it hard for the caretakers to handle the person.[46]

#### 1.1.3. Late Stage
Towards this stage, an AD patient becomes almost dependent on the caretakers. Even though the person cannot talk normally, he can still communicate through signals and signs. In this stage, the person completely forgets how to do things on his own. The person will be unable to feed himself and may also harm himself by making wounds.[60] AD ultimately leads to organ failure and other physical injuries. [65]

### 1.2. Alzheimer's Disease Diagnosis
Biomarker tests, neurophysiological tests, and genetic tests are used to diagnose AD. Alzheimer's disease research is labour-intensive because of the difficulty in tracking behavioural changes over time.[49] However, clinical settings do not make use of biomarker tests. The Mini-Mental State Examination (MMSE) is administered to get a useful clinical result about the patient's mental state. Several aspects of a person's cognitive abilities and memory are evaluated in this test, including how those abilities change over time, how well they can remember things, and how good their short-term memory is. The doctor will ask a series of questions designed to assess the person's common mental abilities. The maximum possible Mini-Mental State Examination (MMSE) score is 30. Both the patient's age and MMSE score are considered. The field of medicine makes extensive use of machine learning for various diagnoses. To choose the best course of action for a patient, a "decision tree" is used.[47] A decision tree is a graphical representation of the various paths one might take when weighing available options. A supervised machine-learning model can be used to find the root of a patient's illness. The supervised learning model is employed to make decisions and conduct an in-depth analysis of a given situation. The most effective method of disease detection and how it stacks up against other patients have both been the subject of studies. Research in the medical field has helped a lot to learn about Alzheimer's disease and make the world a better place to live.[59] Artificial intelligence techniques, as well as long-term memory monitoring, are applied in medical diagnostic technologies.[58]

## 2. Biomarkers in AD Diagnosis

Different biomarkers have been used to diagnose AD, as shown in Figure 2. Tests for amyloid- (A) may reveal the early stages of Alzheimer's disease.[56] The primary biomarker for Alzheimer's disease pathogenesis is the amyloidal precursor protein gene in the long term.[51] Scientists have discovered that the fourth variant of the apolipoprotein E gene on chromosome 14 is responsible for the increased prevalence of Alzheimer's disease in certain families and populations.[52] It was also assumed that elevated blood pressure caused Alzheimer's disease.[53] Impaired nerve growth factor precursor protein (proNGF) has been shown to impact advances in the field of AD diagnosis significantly.[54] The cognitively normal subjects' linguistic performance was used as a biomarker to foretell the onset of AD.[5] A new method based on Cerebrospinal Fluid (fluid in direct contact with the human brain) as a biomarker to diagnose AD through near-infrared (NIR) Raman spectroscopy combined with machine learning analysis was used developed.[6] The expression profiles of genes in the blood are taken as biomarkers, as stated in [7]. For the past several years, neuroimaging biomarkers like MRI and PET have become indispensable in the diagnosis of AD. The functional and anatomical brain changes related to AD are more easily understood from Magnetic Resonance Images (MRI).[8] Many researchers have developed several techniques which use AI algorithms and data sets having MRI images describing the structural changes in the brain, such as cortical and subcortical atrophy, changes in the total brain volume [9], entorhinal cortex and hippocampus atrophy [10] as these are the target locations of neurofibrillary tangles. Some of these methods employed whole brain MRI images as their input [2], whereas others used Region of Interest (ROI) based images [11], Voxel-based images, Slice-based images, and Patch-based images.

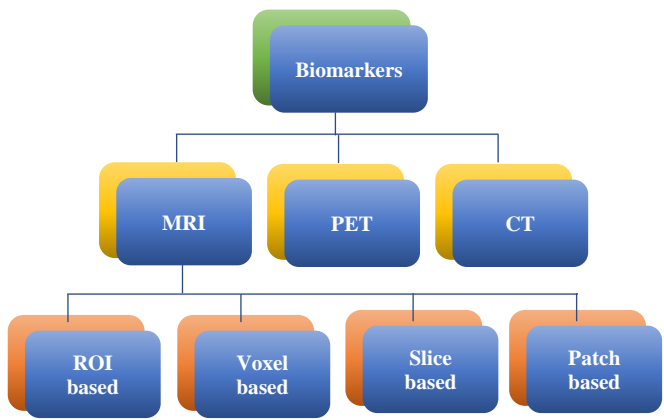

**Fig. 2 Different Biomarkers and feature extraction methods in AD diagnosis**

## 3. Input Data

### 3.1. ROI-based Feature Extraction

The hippocampus, medial temporal lobe, amygdala, and parahippocampal gyrus are just some of the brain regions that begin to lose function and structure in the early stages of Alzheimer's disease. In the study [13], 45 derived ROIs and 2 combined ROIs were used to train and test the model. The Hippocampus, Para hippocampal gyrus, and medial temporal lobe were found to give the best results. Due to the use of 2-D compartmental feature extraction, it was hard to get information from small ROIs with this method. Using 3-D feature extraction methods in ROIs could solve this problem. In [14], the ROI analysis was done on both white matter and grey matter.

### 3.2. Voxel-based Feature Extraction

In Voxel-based feature extraction, the white or grey matter in an MRI is used to figure out the voxel intensity values. The extraction of grey matter, also known as tissue segmentation, is performed on MRI images. In [15], 649 voxel-based morphometry (VBM) images were used to come up with a new framework for voxel-based feature detection. The limitations of ROI-based methods were eased in this method, but the effects of high-dimensional data were still strong. The Extreme Learning Model (ELM), which was made in [16], used features of the VBM. This study extracted the VBM features using a simple feature selection method. This method could be made better by using other methods and large datasets. The brain was first divided into deliberate regions based on its anatomical structure, and then complex non-linear relationships among the voxels were derived [17]. Here, the voxel intensities represent the volumetric measurements of each region.

### 3.3. Slice-based Feature Extraction

In Slice-Based Feature Extraction, 2D features are taken from 3D volumetric brain data to make the 3D features that are needed. The number of hyperparameters is cut down to reduce the number of features. Some of the studies employing 2D image slices consider the standard projections of neuroimages, such as a sagittal (medial) plane, frontal (coronal) plane, or axial (horizontal) plane. The slices that were sampled every 3.0 mm in the coronal direction were used in [18] to make sure that the features were different. To improve classification accuracy, this method could employ better ways to choose which slices to use. In [19], each 3D MRI image is broken up into 2D slices, and an RNN is used after a CNN. Think about how sequences of images relate to each other so that a decision can be made based on all of the slices instead of just one at a time. It was concluded in this study that the accuracy could be improved by combining the 2D image slices with other features, such as memory tests. In [20], a deep learning framework based on Structural MRI (MRI) grey matter slice and an attention mechanism are implemented. This makes the diagnosis of AD more accurate.

### *3.4. Patch-based Feature Extraction*

Patching means selecting arbitrarily shaped regions on an image. This lets to pick up patterns that are related to the disease from MRI. The whole image of the brain is broken up into small areas. Several 3D patches are taken from each area and put into clusters using the K-Means clustering method. [21] A Densenet is then used to learn the patch features for each cluster, which are then grouped for classification. Even though it is not easy to picture the learned features for classification in this method, it can be made better by combining the multimodal brain image analysis. In [22], more than one individual classifier is built for different parts of local patches. These are then put together to make a more accurate classification. This method can be extended to various other modalities to improve the accuracy further.

## 4. Materials and Methods

The most common DL models employed in diagnosing AD are discussed in this article. These include Deep Neural Networks (DNN), Convolutional Neural Networks (CNN), Recurrent Neural Networks (RNN), and Restricted Boltzmann Machines (RBM). More research works have been done on the Deep Learning techniques used for AD diagnosis, as shown in Figure 3.

### *4.1. Deep Neural Network (DNN)*

DNN is a selective and yielding neural network having the same configuration as a Multi-Layer Perceptron (MLP) with more than two layers. DNNs generally consist of one input layer, one output layer, and more than one hidden layer. They are mostly used for classification and regression as DNNs deal with unlabeled and unstructured data. DNNs use the feed-forward characteristic of neural networks for processing data. A popular DNN-based model, LeNet, employed the concatenation of MaxPooling and MinPooling layers in [23]. The DNN model using gadolinium material in [24] secured more accuracy than the machine learning algorithm. DNN using values of AD-related genes obtained from blood samples showed AUC (Area Under Cover) values greater than 0.80 in [7]. The authors of [25] expanded the DNN model by including more hidden units in each layer and adjusting the learning rate hyperparameter to enhance the learning algorithm. According to the findings, the best DNN model for Alzheimer's disease classification uses 11 layers of neurons, with the number of neurons decreasing with each layer.

### *4.2. Convolutional Neural Network (CNN)*

CNNs belong to the family of Deep Neural Networks and are stimulated by the layers of the brain's visual cortex [26]. CNNs are better than ANNs because they can do both feature extraction and classification simultaneously. In the past few years, CNN has become very popular when it comes to using images for things like diagnosing diseases with medical images. The main parts of convolutional neural networks are the input, feature extraction, and classification layers. The feature extraction layers are made up of iterations of a Convolution layer and then a Pooling layer. The Convolutional layer consists of filters and activation maps/layers. The fully connected layers compute the class scores. The study in [27] aimed to inspect the segmentation effects of the CNN model on MRI for Alzheimer's diagnosis and nursing. An ensemble of three deep CNNs following a dense connectivity pattern in [28] achieved more accuracy compared to other baseline deep CNNs. A multimodal unified framework using CNN is proposed in [29]. This study used a fusion of neuroimaging and genetic data for joint classification, and CNN was employed for learning features from MRI, PET, and SNP images. Some studies employed 2D CNNs, whereas others used 3D CNNs for neuroimaging. These studies used 2D CNNs with three convolutional layers for a single-modality approach to finding AD.[30] A multimodal framework for classifying AD used 3D CNN features, gray matter density from MRI, and intensity values from PET scans. [63] In [32], CNN, RNN, and LSTM models have been put together to make a more accurate model.

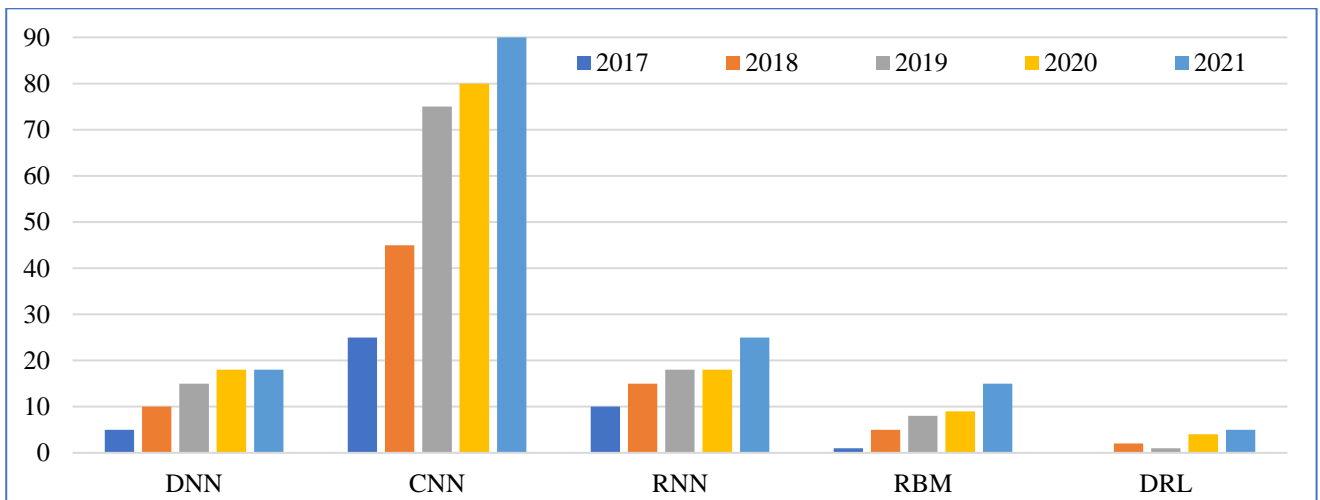

**Fig. 3 Research results published during the last 5 years stating the use of DL techniques in the diagnosis of AD.**

## 4.3. Recurrent Neural Network (RNN)

The Neural Network used in natural language processing and speech recognition is RNN. A relation exists between the input, output, and hidden layers in RNN. The input of a step is the output from the previous step, and the same weights are shared across all layers. The hidden layers in RNN can learn sequences, which helps in the processing of sequential information.

In [33], an RNN model is suggested to fill in missing data during training and testing based on the individual's clinical diagnosis, cognitive state, and ventricular volume each month into the future. Convolutional neural networks (CNNs) and recurrent neural networks (RNNs) are proposed in [34] as part of a model for detecting and classifying Alzheimer's disease. The spatial properties of MR images are learned using a Convolutional neural network (CNN).

The outputs of the CNN are used as inputs to a recurrent neural network (RNN) comprised of three cascaded bidirectional gated recurrent units (BGRU) that extract longitudinal characteristics. [35] suggests a hybrid convolutional and recurrent neural network using structural MRI data for a more in-depth look at the hippocampus. In this research, the RNN is cascaded to learn the more advanced features of AD categorization by combining the information from the left and right hippocampus.

In [36], an improved many-to-one LSTM RNN model with one input layer, one output layer, and two hidden layers verifies that RNN can efficiently handle the AD progression issue by fully capitalizing on the intrinsic temporal and medical patterns obtained from the patient's past appointments. Using a data-driven technique and RNN to classify AD, a new feature sequence representation was developed [37].

## 4.4. Restricted Boltzmann Machine (RBM)

RBMs are a kind of random and generative neural network with both an exposed and a concealed layer. All of the input layer neurons in this design communicate with their hidden layer counterparts, but they do not interact with their peers in the same layer. Research has shown that RBMs can learn internal representations. Hence they have been employed in diagnosing AD.

A collection of RBMs (also called Deep Belief Networks, DBN) for feature extraction and SVM (Support Vector Machine) is employed for classification.[38] The RBMs used in this work underwent unsupervised pre-training. For AD detection, researchers turn to the DBN [39], a graphical model for obtaining a deep hierarchical representation of training data.

In [40], a novel method for a high-level latent and shared feature representation from neuroimaging modalities using the Deep Boltzmann Machine (a deep network with RBM as a building block) is implemented. In this study, DBM is utilized to uncover the 3D patch's latent hierarchical feature representation, and a technique for systematic joint feature representation between MRI and PET patch pairs is developed.

## 4.5. Deep Reinforcement Learning (DRL)

DRL combines the features of Deep Learning (DL) and Reinforcement Learning (RL). RL is a branch of ML in which agents are employed to study about the environment, and the agents learn by receiving rewards from the environment.

In [61], the proposed approach produced more accurate results when compared with some existing forms of Reinforcement Learning models. This system in [61], based on reinforcement learning and neural networks, could generate imbalanced segment classes of datasets.

A framework combining Differential Equations (DE) and RL for modelling AD progression is depicted in [62]. The model was found to be better at producing 10-year cognition trajectories for AD than the existing models.

## 5. Results and Discussion

Different ML approaches have been used for AD classification, and we compared some of them in this study. In Table 1, we can see how each ML technique performed in comparison to the others. While analyzing the comparison of various DL techniques used for AD classification, it was found that while using the DNN technique, Lenet with ADNI dataset has an accuracy of 96.64%, whereas variant DNN with 20 hidden layers using OASIS dataset has an accuracy of 91.00% and Feed Forward DNN with ADNI dataset has 79.3% accuracy. While focusing on the second technique CNN, variant Deep CNN with MRI and dataset Oasis was found to have 93% accuracy. Secondly, while focusing on variant supervised 2-D CNN, with multi-modality MRI, PET, and SNP with dataset ADNI has an accuracy of 98.22%. Examining a variant ensemble of CNN-RNN-LSTM with an MRI data set of ADNI yields an accuracy of 92.22%. RNN with 3 cascaded BGRU using MRI dataset of ADNI has an accuracy of 91.33%, whereas LSTM RNN with multi-modality CDR, GDS, FAQ dataset of National Alzheimer's coordinating center has an accuracy of 99%. The Area Under Curve (AUC) is achieved more in CNN when compared to DNN and RNN. Deep Reinforcement Learning has shown more accuracy than RL with the OASIS dataset. Moreover, the multi-modality approaches have shown better results than the single-modality approaches.

Table 1. Comparison of various ML techniques used for AD classification

| Technique | Variants | Modality | Dataset Source | AUC | Accuracy |
|---|---|---|---|---|---|
| DNN | DNN with two hidden layers | AD-related genes from blood samples | ADNI, ANM1, ANM2 | 0.80 | |
| | LeNet | MRI | ADNI | | 96.64% |
| | DNN with 20 hidden layers | MRI | OASIS | | 91.00% |
| | Feed-Forward DNN | MRI | ADNI | | 79.3% |
| CNN | Improved CNN | MRI | IGMC hospital | | |
| | Deep CNN | MRI | OASIS | | 93% |
| | Supervised 2D CNN | MRI, PET, SNP | ADNI | | 98.22% |
| | CNN | fMRI | Huashan Hospital of Fudan University | | 95.59% |
| | Deep 3D CNN | MRI, PET | ADNI | 0.92 | |
| | Ensemble of CNN-RNN-LSTM | MRI | ADNI | | 92.22% |
| RNN | Minimal RNN | MRI, PET, CSF measures | ADNI | | |
| | RNN with 3 cascaded BGRU | MRI | ADNI | | 91.33% |
| | Hybrid RNN with CNN model | MRI | ADNI | 0.91 | |
| | LSTM RNN | CDR, GDS, FAQ | National Alzheimer's Coordinating Centre | | 99% |
| | Data-Driven RNN | Speech and Neuropsychological test | | 0.83 | |
| RBM | Stacked RBM | MRI, PET | ADNI | | 91.4% |
| | DBN | MRI | OASIS | | 91.76% |
| | Multimodal DBM | MRI, PET | ADNI | | 95.35% |
| DRL | RL (Reinforcement Learning) | MRI | OASIS | | 82.23% |
| | DRL (Deep Reinforcement Learning) | MRI | OASIS | | 84.34% |
| | DRL-XGBOOST | MRI | OASIS | | 90.23% |

## 6. Conclusion

Alzheimer's is a terrible disease that affects a person's mind and behavior. Because of this, it is important to diagnose it early. In the past few years, researchers have tried out several Deep Learning algorithms and methods to detect AD—most of the studies that were looked at used CNN to put things into groups. In 90% of the research, ADNI is the main data set. The ADNI is a multicenter longitudinal investigation that looks for clinical, imaging, genetic, and biochemical signs of AD for its early diagnosis. Using sophisticated deep learning algorithms, these datasets may enable improved and earlier prediction of AD via the identification of the best mix of diverse biomarkers. Alzheimer's disease (AD) is one of the most common types of dementia today. World Alzheimer's Report (2018) says this disease affected about 50 million people in 2018. This number is expected to triple by 2050. Mild cognitive impairment (MCI) is the initial stage of Alzheimer's disease progression. However, not everyone with MCI develops Alzheimer's disease. Understanding the transformations that lead from MCI to AD is, thus, a primary focus of current studies. These shifts may be measured using medical imaging or other techniques like blood plasma spectroscopy. The availability of several open-source datasets for Alzheimer's research has accelerated the field. ADNI (adni.loni.usc.edu), AIBL (aibl.csiro.au), and OASIS are only a few of the most widely used databases (oasis.csiro.au). Open to the public for the first time. The J-ADNI database contains data from longitudinal studies conducted in Japan.

Furthermore, MRI image processing is quite labor-intensive. The Wellcome Centre for Human Neuroimaging has released free tools, such as Statistical Parametric Mapping (SPM), to facilitate the analysis of MRI scans. To perform voxel-based morphometry (VBM) using MRI data, SPM is used. Many scientists make use of Freesurfer, a widely-used open-source application designed for volume-based morphometry. In the last ten years, machine learning techniques have been found to be very useful for diagnosing Alzheimer's. Common categorisation methods include Support Vector Machines (SVMs), artificial neural networks, and deep learning. The way the optimization problem is set up is the main difference between SVM and ANN. Both SVM and ANN have an important step called "feature extraction".

On the other hand, deep learning builds the step of extracting features into the learning model itself. Deep learning is helpful for big data sets, especially image data. Some researchers also used methods called "ensembles" to improve the accuracy of Alzheimer's classification. DL combined with RL can be used for various MRI processing techniques. Very few studies have employed Deep Reinforcement Learning for AD prediction. DRL has shown better prediction results when compared with RL. Hence the scope of more studies exists in DRL-based Alzheimer's diagnosis.